# Integrated Laser Scanning and Image-Based Topology Optimization Techniques for Detection and Quantification of Visible and Subsurface Structural Defects

Mehrdad S. Dizaji[1] and Devin K. Harris[2]

[1] Department of Engineering Systems and Environment, University of Virginia, Charlottesville, VA, USA; ms4qg@virginia.edu
[2] Department of Engineering Systems and Environment, University of Virginia, Charlottesville, VA, USA; dharris@virginia.edu

## ABSTRACT

Reliable characterization of structural defects requires methods capable of resolving both directly observable surface damage and damage that is not visible from the inspected surface. This study presents two complementary non-contact, vision-based approaches for the detection and quantitative characterization of defects in structural components. The first approach employs high-resolution laser scanning to generate three-dimensional (3D) point clouds of damaged steel specimens. Comparative processing of measured and reference point clouds is used to localize damaged regions, quantify geometric loss, and transfer the measured defect geometry to a finite element representation. The second approach combines full-field surface deformation measurements obtained using three-dimensional digital image correlation (3D-DIC) with finite element model updating and topology optimization. In this inverse framework, measured surface response is used to infer subsurface abnormalities through their influence on the spatial distribution of structural response. Experimental steel-beam specimens containing controlled smooth defects and randomly distributed defects are used to evaluate the approaches. Comparisons with milling-based ground-truth measurements demonstrate that both methods can identify and quantify defect geometry, while providing complementary information for visible and subsurface damage assessment. The combined framework establishes a pathway toward high-fidelity, non-contact structural condition assessment and model updating for components with complex and irregular damage.

*Keywords: laser scanning; digital image correlation; topology optimization; structural identification; finite element model updating; damage detection; nondestructive evaluation.*

## 1. INTRODUCTION AND METHODOLOGICAL FRAMEWORK

Structural defects frequently exhibit irregular and stochastic geometries that are difficult to represent using conventional idealized numerical models. Accurate characterization of such defects is important because local geometry and material-property variations can significantly influence structural response. This study therefore investigates a measurement-driven framework for generating high-fidelity representations of structural damage using two complementary vision-based techniques. The first technique targets visible surface damage through high-resolution laser scanning and 3D point-cloud processing. The second targets non-visible or subsurface abnormalities by coupling full-field 3D-DIC measurements with finite element model updating (FEMU) and topology optimization (TO). In the combined framework, regions of visible damage identified from surface measurements can inform the numerical model, while deformation fields measured during loading provide additional information for identifying internal abnormalities. The overall objective is to detect, localize, and quantify structural defects while improving the correspondence between measured behavior and numerical simulation.

## 2. LASER-SCANNING-BASED SURFACE DAMAGE CHARACTERIZATION

The first approach uses direct 3D geometric measurements obtained by high-resolution laser scanning to characterize surface defects and update the geometry of the corresponding numerical model (Figure 1).

Depending on the damage mechanism, the measured geometry can be represented through localized element removal or reduction, introduction of geometric imperfections, or loss of connectivity. In the present study, steel beams were scanned to capture both nominal geometry and defect-induced surface variations (Figure 2). Following the procedure developed in prior work, the measured point-cloud data were compared with an as-built or reference point cloud to identify localized geometric deviations associated with damage. Computer-vision-based processing was then used to locate, isolate, and quantify the damaged regions. Rather than reconstructing the complete surface of the structural component, only point-cloud subsets associated with the detected defects were extracted and mapped to the existing solid model. The locally modified model was subsequently remeshed for finite element analysis. This localized updating strategy preserves the spatial resolution available from the point cloud in damaged regions while limiting unnecessary computational expense elsewhere in the model (Figure 3).

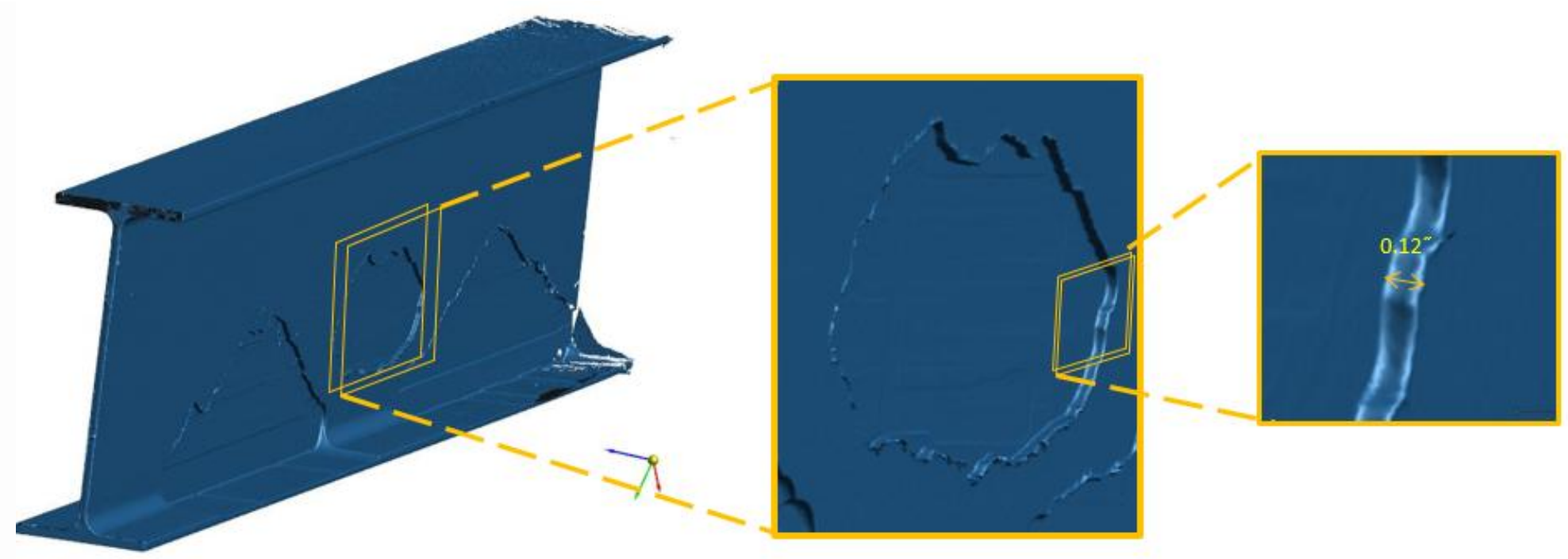

*Figure 1. Surface-scan data collection for the 4-ft steel beam containing randomly distributed defects on the rear surface.*

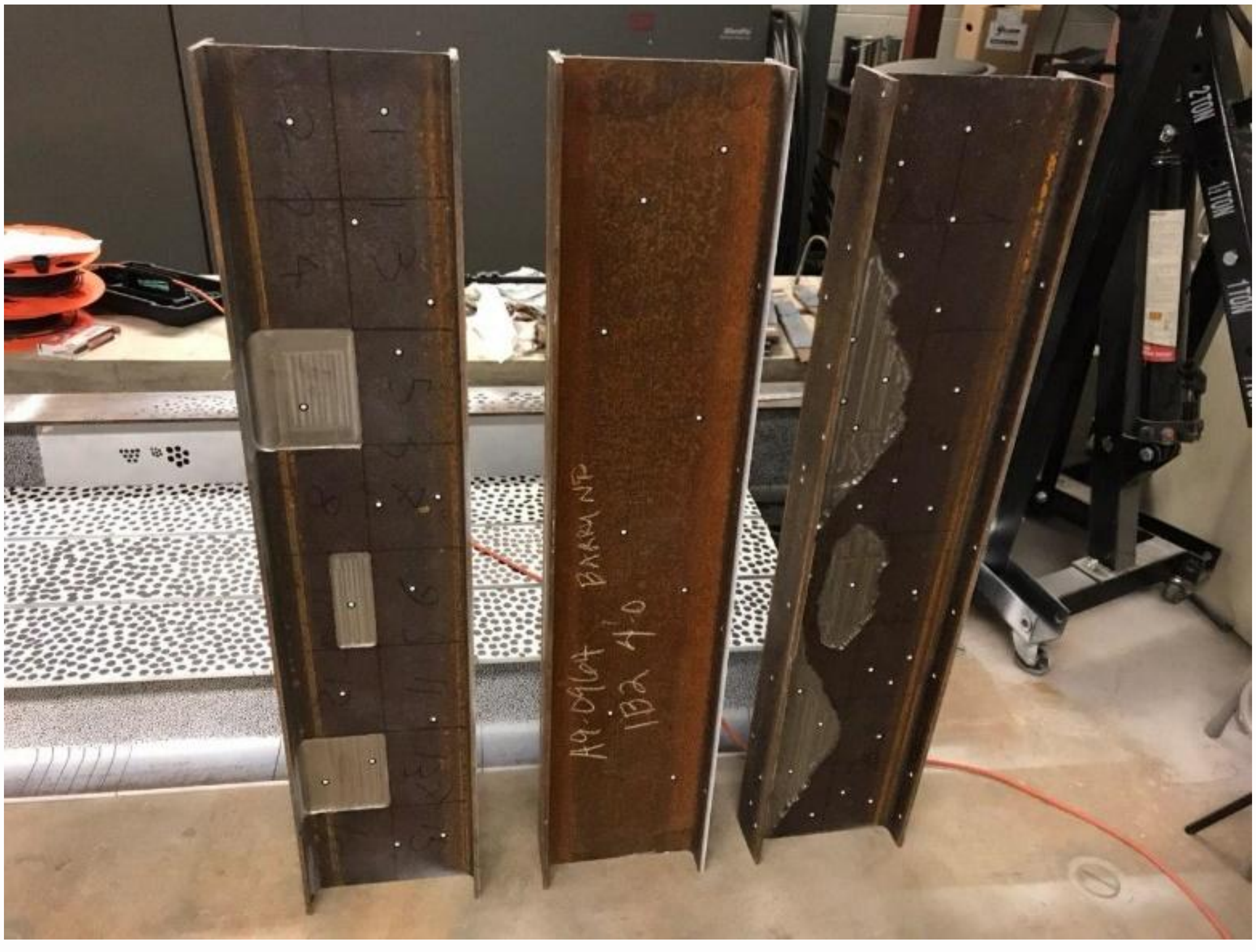

*Figure 2. Steel-beam specimens used in the experimental investigation.*

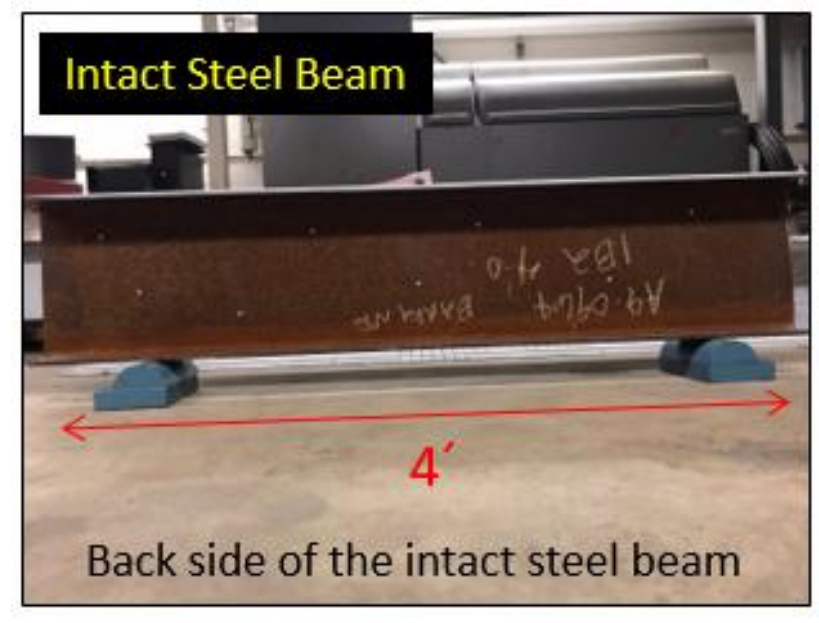
Intact Steel Beam
4′
Back side of the intact steel beam
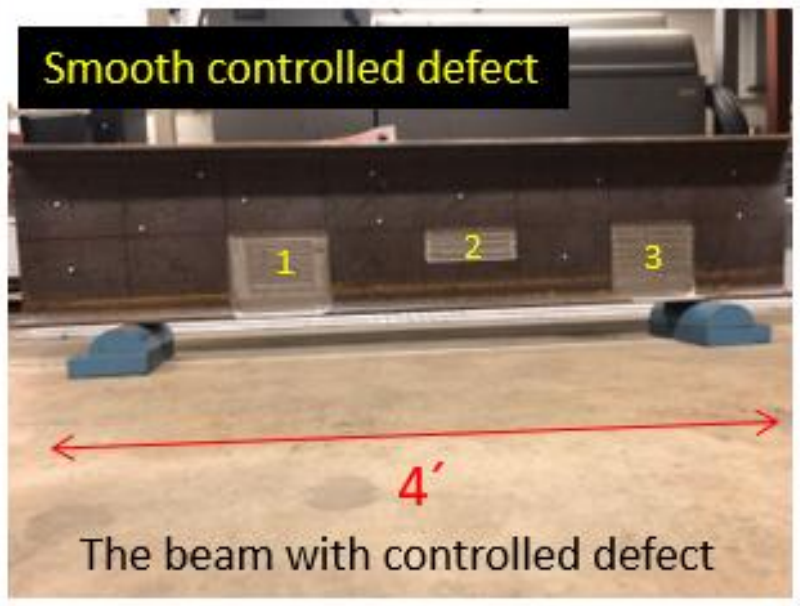
Smooth controlled defect
1
2
3
4′
The beam with controlled defect
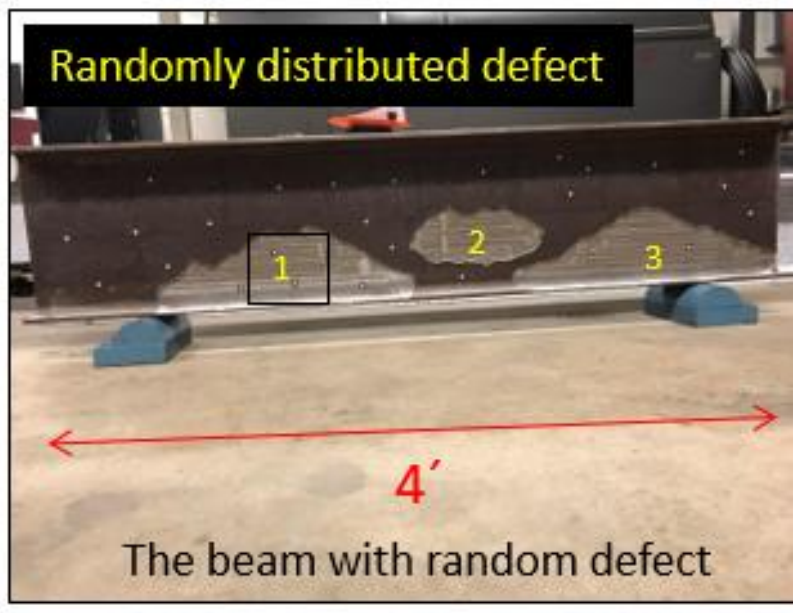
Randomly distributed defect
1
2
3
4′
The beam with random defect
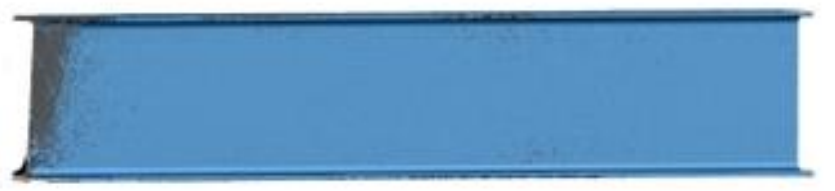
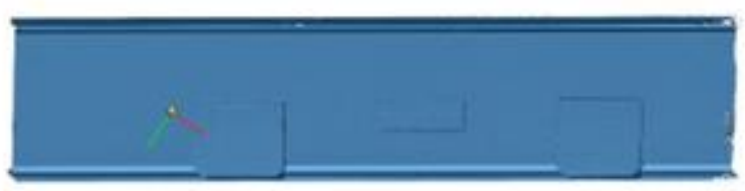
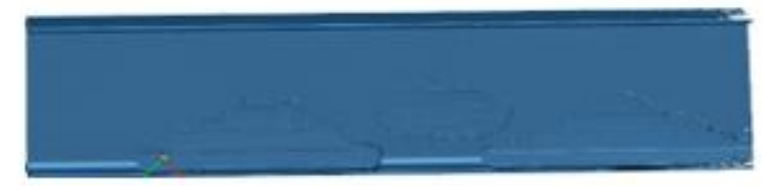
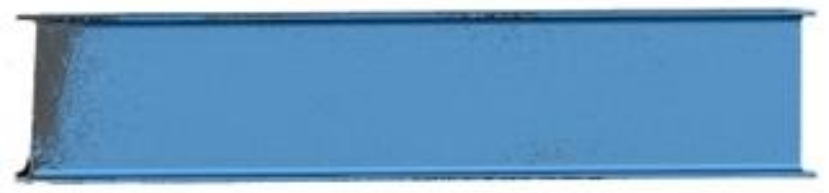
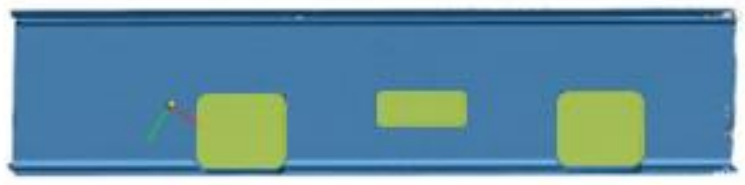
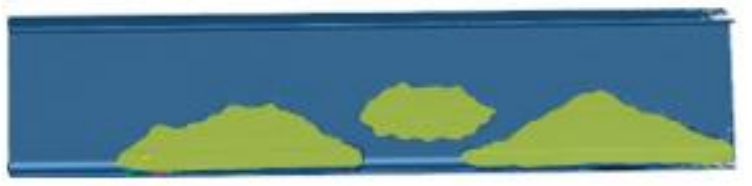

(a)

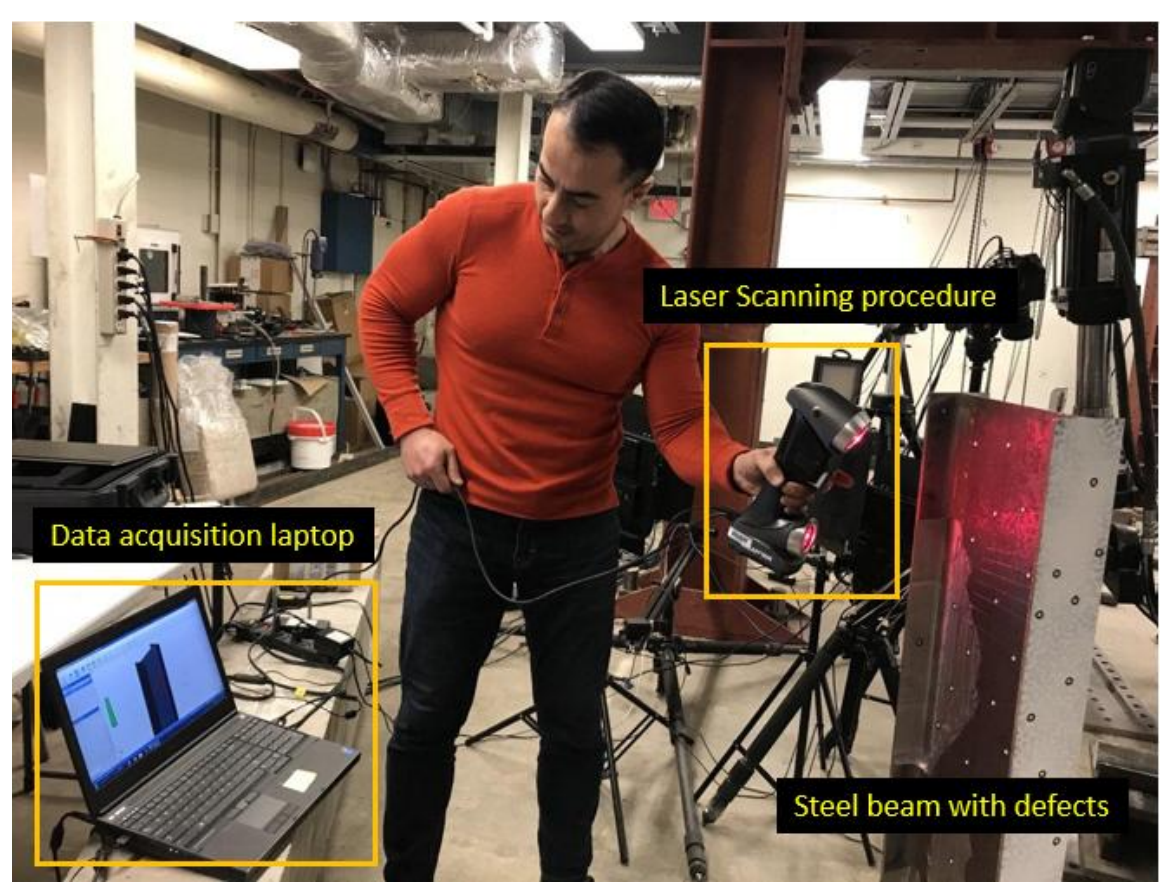
Laser Scanning procedure
Data acquisition laptop
Steel beam with defects

(b)

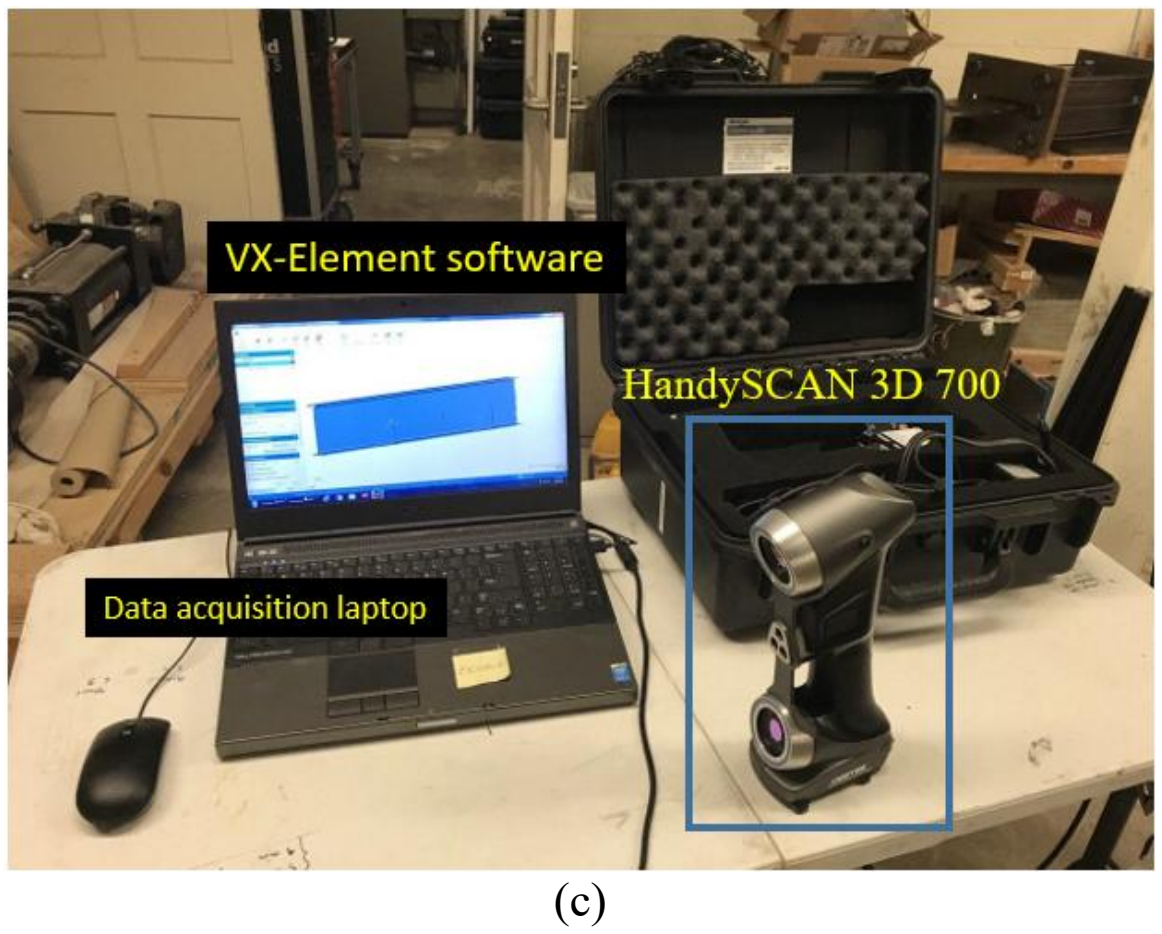
VX-Element software
HandySCAN 3D 700
Data acquisition laptop

(c)

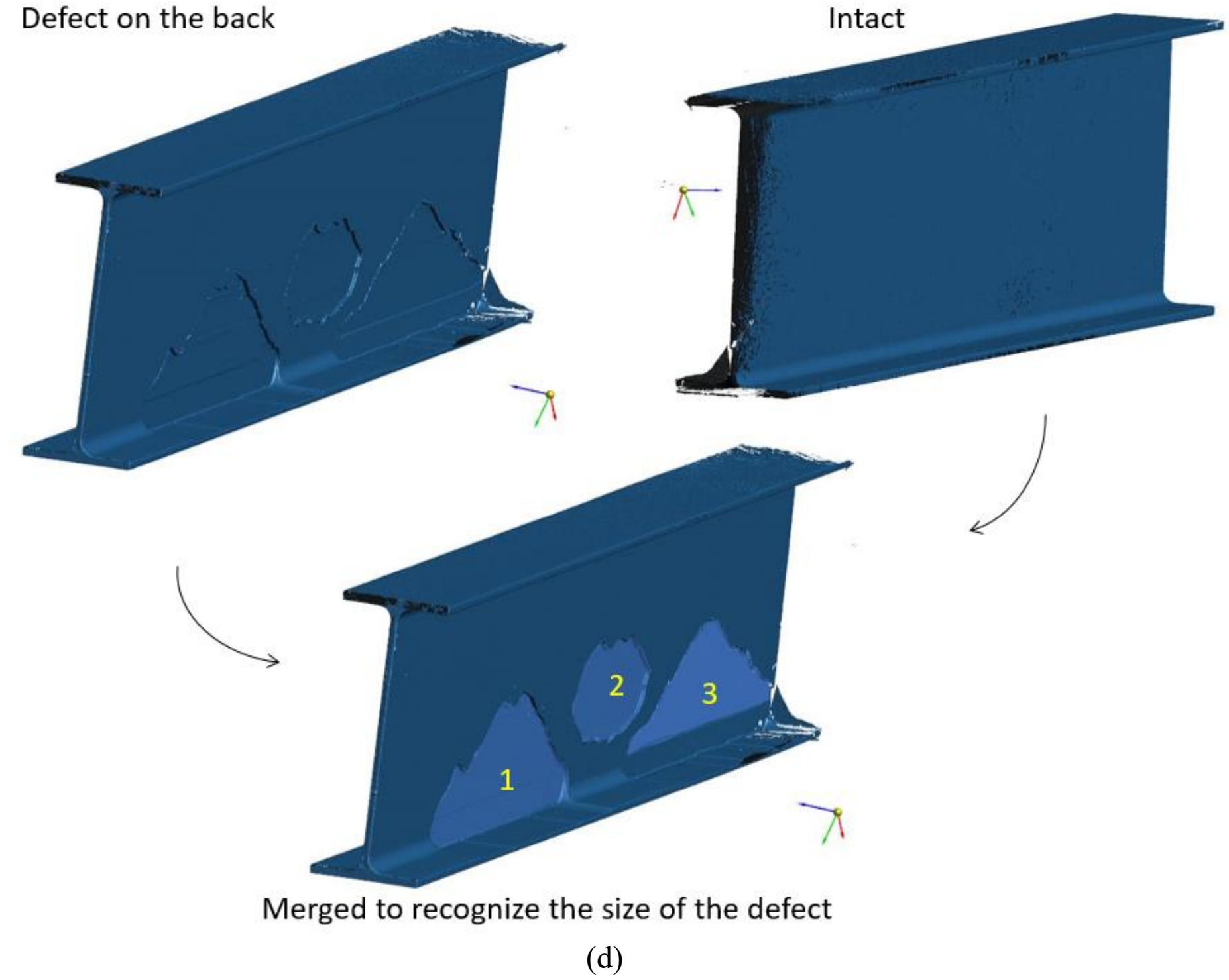

(d)

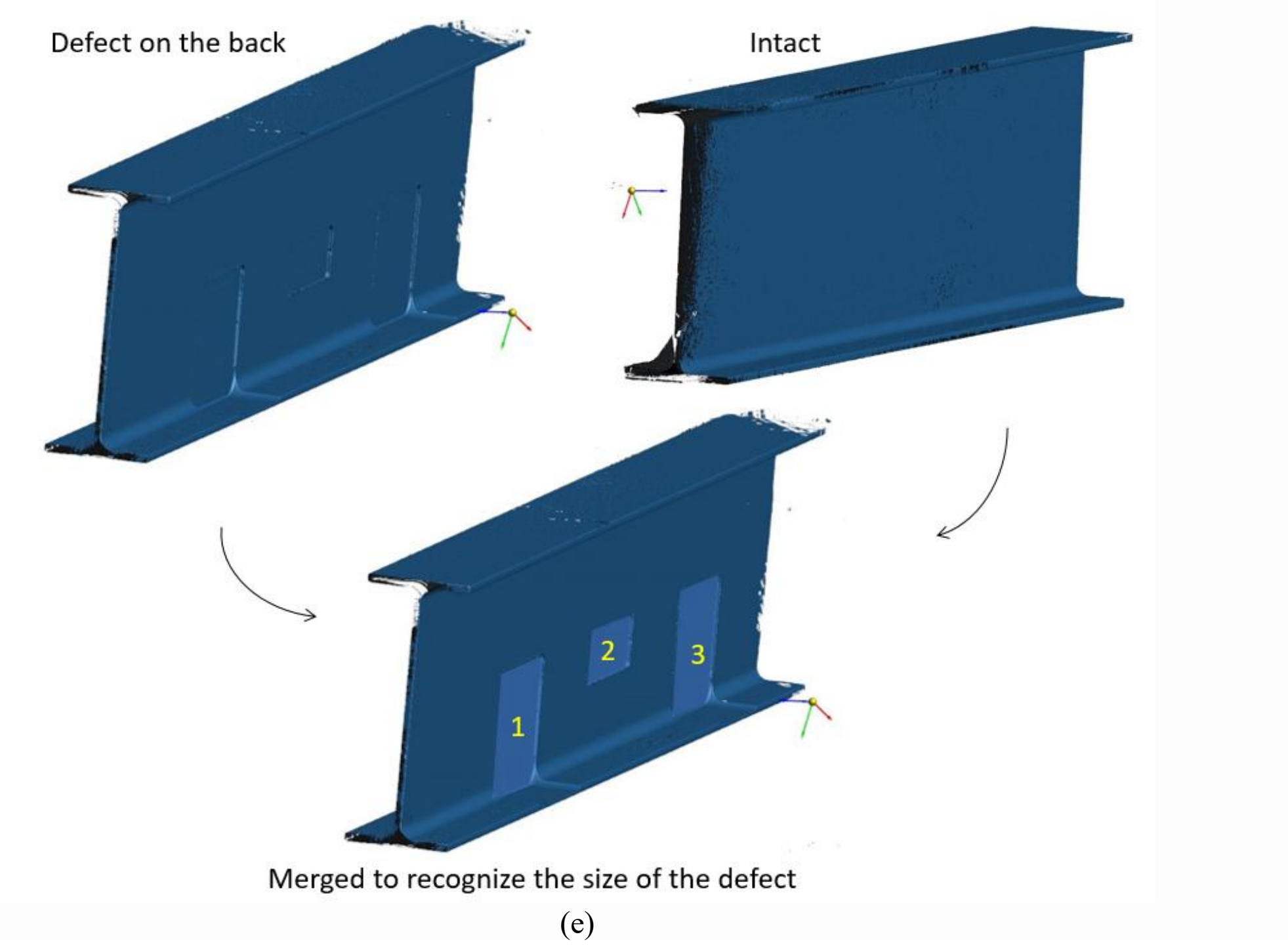

(e)

*Figure 3. Laser-scanning equipment and experimental setup: (a) scanning procedure; (b) laser-scanning device; and steel-beam specimens comprising an intact beam, a beam with controlled smooth defects on the rear surface, and a beam with randomly distributed defects on the rear surface.*

## 3. 3D-DIC AND TOPOLOGY-OPTIMIZATION-BASED SUBSURFACE DAMAGE IDENTIFICATION

The second approach is based on the hypothesis that internal abnormalities can be inferred from their influence on measured surface deformation fields. Subsurface defects may be represented as localized anomalies in material-property distributions or as geometric discontinuities that alter the structural response. Accordingly, spatial perturbations in full-field displacement and strain measurements can provide indirect information regarding the location, shape, and severity of otherwise unobservable damage. The proposed framework combines full-field deformation measurements from 3D digital image correlation (3D-DIC) with analogous response fields predicted by a finite element model. These data are incorporated into an inverse structural-identification procedure formulated within a topology-optimization framework. During model updating, candidate finite elements in regions associated with potential damage are adjusted to reduce discrepancies between experimentally measured and numerically predicted response fields. The approach can therefore incorporate visible-damage information obtained from laser scanning while using static or dynamic 3D-DIC measurements to identify additional non-visible abnormalities. Structural steel specimens provide a nominally homogeneous material system for proof-of-concept validation and facilitate evaluation of the method under controlled defect configurations. The framework builds on the authors' prior work in full-field structural identification and subsurface damage detection [1–7] and extends those concepts toward combined surface and internal defect characterization (Figure 4).

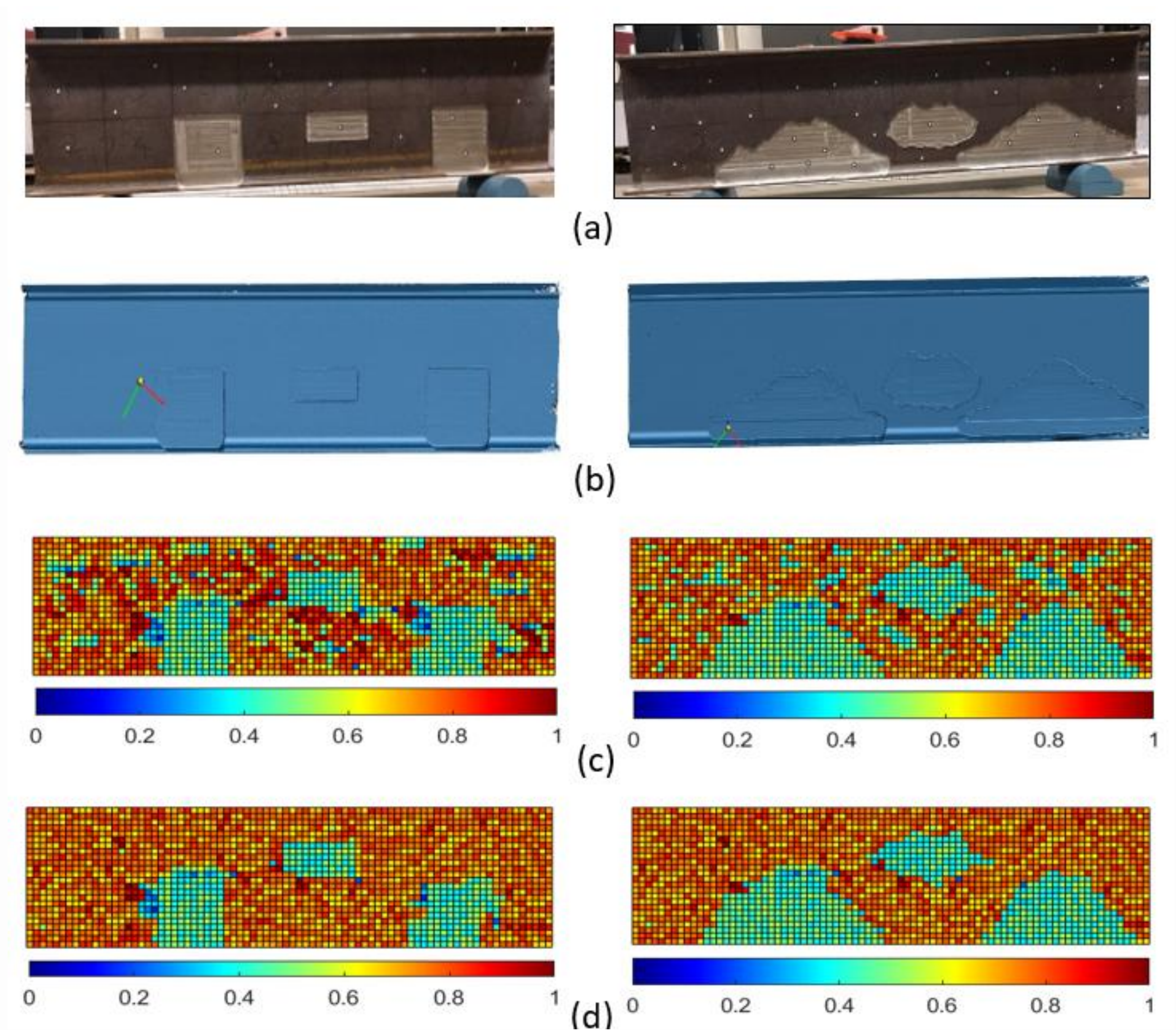

*Figure 4. Comparison of laser-scanning-based defect detection and topology-optimization results: (a) 4-ft steel beam with smoothly distributed controlled defects on the rear surface; (b) 4-ft steel beam with randomly distributed defects on the rear surface; (c) topology-optimization result before post-processing; and (d) topology-optimization result after post-processing.*

## 4. RESULTS AND DISCUSSION

### 4.1. Controlled Smooth Defects

Table 1 presents the defect-volume measurements for the steel beam containing three controlled smooth defects. The milling CAD model was used as the geometric reference, with defect volumes of 2.88, 1.44, and 2.88 for Defects 1, 2, and 3, respectively. Visual inspection yielded corresponding estimates of 3.00, 1.52, and 3.05; laser scanning produced 3.12, 1.77, and 3.10; and topology optimization produced 3.15, 1.98, and 3.19. These results show that all three measurement strategies captured the overall magnitude of the prescribed damage, although the level of agreement varied among defects and methods.

For Defect 1, the visual-inspection estimate differed from the milling CAD reference by approximately 4.2%, compared with approximately 8.3% for laser scanning and 9.4% for topology optimization. A similar pattern was observed for Defect 3, for which the corresponding absolute percentage differences were approximately 5.9%, 7.6%, and 10.8%. Thus, for the two larger smooth defects, all three methods provided relatively consistent estimates, with visual inspection showing the closest numerical agreement with the prescribed geometry.

The largest discrepancies occurred for Defect 2, the smallest controlled defect. Relative to the reference value of 1.44, visual inspection produced an estimate of 1.52 (approximately 5.6% difference), whereas laser scanning and topology optimization produced estimates of 1.77 and 1.98 (approximately 22.9% and 37.5% differences, respectively). Because Defect 2 has a substantially smaller reference volume, modest absolute differences translate into comparatively large percentage errors. The result also suggests greater sensitivity of small-defect reconstruction to spatial measurement resolution and to the discretization and post-processing procedures used to delineate the damaged region.

Across all three controlled smooth defects, the mean absolute percentage deviations from the milling CAD reference were approximately 5.2% for visual inspection, 13.0% for laser scanning, and 19.2% for topology optimization. Although visual inspection provided the closest quantitative agreement for this controlled configuration, the significance of the topology-optimization result lies in its different measurement principle: the defect is inferred from its influence on the measured structural response rather than solely from direct geometric observation. This distinction supports the use of the inverse approach for damage that may be inaccessible or not directly visible from the inspected surface.

*Table 1. Defect-volume measurements obtained using different methods for the 4-ft steel beam with controlled smooth defects on the rear surface.*

| **Method** | **Defect Index** | | |
|---|---|---|---|
| | **1** | **2** | **3** |
| Milling CAD machine ($in.^3$) | 2.88 | 1.44 | 2.88 |
| Visual Inspection ($in.^3$) | 3.00 | 1.52 | 3.05 |
| Laser Scanner ($in.^3$) | 3.12 | 1.77 | 3.10 |
| Topology Optimization ($in.^3$) | 3.15 | 1.98 | 3.19 |

### 4.2. Randomly Distributed Defects

Table 2 summarizes the measurements for the beam containing randomly distributed defects. The milling CAD reference volumes for Defects 1, 2, and 3 were 4.94, 2.29, and 5.06, respectively. Visual inspection produced estimates of 4.31, 2.06, and 4.50; laser scanning produced 4.61, 2.22, and 5.01; and topology optimization produced 4.77, 2.48, and 5.22. Compared with the smooth-defect specimen, the non-contact methods exhibited notably stronger overall agreement with the prescribed defect volumes for this irregular damage configuration.

For Defect 1, topology optimization provided the closest estimate, with a value of 4.77 compared with the reference value of 4.94, corresponding to an absolute percentage difference of approximately 3.4%. Laser scanning differed by approximately 6.7%, while visual inspection underestimated the defect volume by approximately 12.8%. For Defect 2, laser scanning provided the closest result: the estimate of 2.22 differed from the reference value of 2.29 by approximately 3.1%, compared with approximately 10.0% for visual inspection and 8.3% for topology optimization.

For Defect 3, laser scanning showed particularly strong agreement with the reference geometry. The laser-scanning estimate of 5.01 was only approximately 1.0% different from the milling CAD value of 5.06. Topology optimization also produced close agreement, with an estimate of 5.22 and a difference of approximately 3.2%. Visual inspection yielded 4.50, corresponding to an approximately 11.1% difference. These results demonstrate that both laser scanning and topology optimization can provide quantitatively useful estimates for complex, irregular defect geometries.

When all three randomly distributed defects are considered, the mean absolute percentage deviations were approximately 11.3% for visual inspection, 3.6% for laser scanning, and 5.0% for topology optimization. Laser scanning therefore provided the highest overall quantitative accuracy for this specimen, with topology optimization following closely. Both non-contact approaches substantially improved upon the manual visual-inspection estimates for the randomly distributed damage configuration.

*Table 2. Defect-volume measurements obtained using different methods for the 4-ft steel beam with randomly distributed defects on the rear surface.*

| Method | Defect Index | | |
|---|---|---|---|
| | 1 | 2 | 3 |
| Milling CAD machine ($in.^3$) | 4.94 | 2.29 | 5.06 |
| Visual Inspection ($in.^3$) | 4.31 | 2.06 | 4.50 |
| Laser Scanner ($in.^3$) | 4.61 | 2.22 | 5.01 |
| Topology Optimization ($in.^3$) | 4.77 | 2.48 | 5.22 |

### 4.3. Comparative Discussion

The results from Tables 1 and 2 demonstrate that measurement performance depends on defect morphology and size as well as on the physical principle underlying each technique. Visual inspection provided the closest estimates for the controlled smooth defects, whereas laser scanning produced the best overall agreement for the randomly distributed defects. The stronger performance of laser scanning for irregular defects is consistent with its ability to represent a three-dimensional surface using dense point-cloud measurements rather than a limited set of manually measured dimensions. This capability is particularly relevant to naturally occurring deterioration, which commonly exhibits nonuniform boundaries and spatially varying depth.

Laser scanning and topology optimization should therefore be regarded as complementary rather than competing approaches. Laser scanning directly characterizes accessible surface geometry and can provide localized geometric information for finite element model updating. In contrast, the 3D-DIC/topology-optimization framework estimates abnormalities indirectly from perturbations in full-field structural response. The approximately 5.0% mean deviation obtained by topology optimization for the randomly distributed defects is particularly encouraging because the reconstruction is derived from response-based inverse identification rather than direct measurement of the defect surface.

The comparison also highlights the influence of defect scale. The relatively large percentage deviations for Defect 2 in Table 1 indicate that smaller defects may be more sensitive to measurement resolution, finite element mesh density, reconstruction thresholds, and post-processing choices. For the inverse identification framework, additional uncertainty may arise from material-property assumptions, boundary-condition representation, measurement noise, and regularization. These factors should be examined systematically in future studies to establish detection limits and uncertainty bounds for quantitative defect reconstruction.

Overall, the experimental comparisons support the feasibility of integrating direct three-dimensional geometric measurements with full-field image-based structural identification. Laser scanning provides high-resolution information for visible surface damage, while 3D-DIC combined with topology optimization offers a pathway for identifying abnormalities through their influence on structural response. Agreement between these independent measurement principles can increase confidence in defect localization and quantification and provides a foundation for an integrated nondestructive evaluation framework applicable to geometrically complex or partially hidden structural damage (Figure 5).

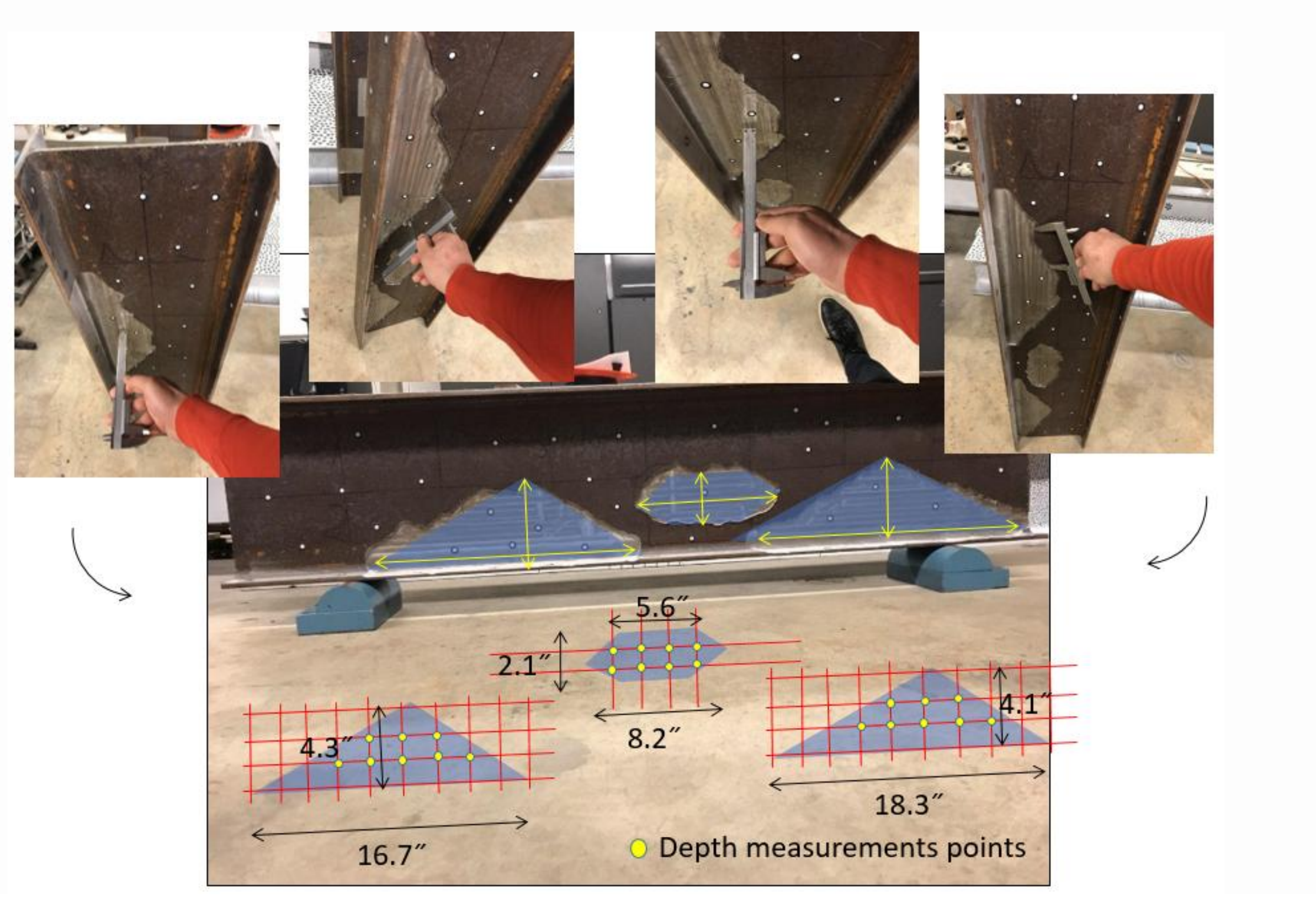

*Figure 5. Visual measurement of defect depth, width, and length.*

## 5. CONCLUSIONS

This study demonstrates a complementary measurement and identification framework for characterizing both visible surface defects and non-visible structural abnormalities. High-resolution laser scanning provides direct geometric information for localizing surface damage, quantifying material loss, and transferring measured defect geometry into finite element models. In parallel, the 3D-DIC/topology-optimization framework uses full-field deformation measurements to infer abnormalities that may not be directly observable from the inspected surface. For the controlled smooth-defect specimen, the mean absolute deviations from the milling-based reference values were approximately 5.2% for visual inspection, 13.0% for laser scanning, and 19.2% for topology optimization. For the randomly distributed defects, the corresponding mean absolute deviations were approximately 11.3%, 3.6%, and 5.0%, respectively. These results indicate that measurement accuracy varies with defect morphology and method; importantly, the two proposed non-contact approaches provide complementary capabilities rather than redundant measurements. Their integration offers a practical basis for measurement-driven finite element model updating and structural condition assessment when damage geometry is complex or partially hidden. Further work should evaluate the framework across a broader range of defect geometries, loading conditions, material systems, and full-scale structural components.